\pdfoutput=1
\documentclass[11pt]{article}

\usepackage{ACL2023}

\usepackage{times}
\usepackage{float}
\usepackage{latexsym}

\usepackage{graphicx}

\usepackage[T1]{fontenc}
\usepackage[utf8]{inputenc}

\usepackage{microtype}

\usepackage{inconsolata}
\usepackage{dirtytalk}
\usepackage{tabularx}

\title{The distribution of discourse relations within and across turns in spontaneous conversation}

\author{S. Magalí López Cortez \qquad Cassandra L. Jacobs \\
  Department of Linguistics \\
  University at Buffalo \\
  Buffalo, NY, USA \\
  \texttt{solmagal;cxjacobs@buffalo.edu} \\}

\begin{document}
\maketitle
\begin{abstract}
Time pressure and topic negotiation may impose constraints on how people leverage discourse relations (DRs) in spontaneous conversational contexts.
In this work, we adapt a system of DRs for written language to spontaneous dialogue using crowdsourced annotations from novice annotators. 
We then test whether discourse relations are used differently across several types of multi-utterance contexts.
We compare the patterns of DR annotation within and across speakers and within and across turns. 
Ultimately, we find that different discourse contexts produce distinct distributions of discourse relations, with single-turn annotations creating the most uncertainty for annotators. Additionally, we find that the discourse relation annotations are of sufficient quality to predict from embeddings of discourse units.
\end{abstract}

\section{Introduction}
Discourse relations (DRs) such as Elaboration, Background and Explanation, hold between discourse units contributing to the coherence of a text. 
Annotation of discourse relations has received attention for its relevance to discourse parsers, with applications in question answering systems (e.g. \citealp{jansen2014discourse}), text summarization (e.g. \citealp{liu-chen-2019-exploiting}), sentiment classification (e.g. \citealp{kraus2019sentiment}), and machine translation (e.g. \citealp{meyer2012using}). 
However, most of the annotated data and systems have focused on written language, with a few exceptions \citep[e.g.,][]{tonelli-etal-2010-annotation,zeldes2017gum,scholman-etal-2022-discogem}.
In spoken dialogue or multiparty conversation, participants must quickly juggle a variety of tasks, such as responding to another person to solve a problem \cite{Levinson2015TimingIT} or negotiating the question under discussion \citep{roberts2012information}, often under considerable time pressure that is less present in written production. 
In addition to these time demands, it is unclear whether spontaneous conversation demonstrates the same patterns of DRs as observed in written language (see \citealp{crible2017discourse}, for a discussion of spoken vs. written use of discourse markers).

Perhaps unsurprisingly, the vast majority of work on discourse relations has focused either on written texts, especially news text \citep{carlson2003building,prasad-etal-2008-penn,prasad-etal-2018-discourse}, or highly structured conversations that are constrained by a particular game \citep{afantenos-etal-2015-discourse,asher-etal-2016-discourse}. 
Some recent corpora contain spoken monologues  \citep{scholman-etal-2022-discogem}, and spoken conversations \citep{tonelli-etal-2010-annotation,zeldes2017gum}, but the field still largely lacks annotated corpora of spontaneous dialogue.

Thus, our goal is to present the first efforts towards an annotated corpus of DRs for spontaneous spoken conversation, with particular attention to relations across different contexts within a conversation. 
We analyze the patterns of DR annotation within and across speakers and within and across turns and test the coherence of annotators' decisions.

\section{Related Work}

Most currently available corpora annotated with DRs have focused on written language or spoken monologues.  An exception is the Georgetown University Multilayer (GUM) corpus \cite{zeldes2017gum}, which has a set of conversations annotated within Rhetorical Structure Theory \citep[RST,][]{mann1987rhetorical}, following the guidelines of the RST Discourse Treebank \citep[RST-DT,][]{carlson2003building}.
But it is an open question whether the DRs that have been identified for news texts are appropriate for conversational data. \citet{tonelli-etal-2010-annotation} adapt the PDTB framework to annotate a subset of a corpus of Italian conversations about software and hardware troubleshooting, and suggest modifications to the framework to account for spoken data.

Discourse relations corpora have usually been annotated by experts, but some recent corpora have been annotated by novice annotators, such as university students, in the case of the GUM corpus \citep{zeldes2017gum}, or crowdsourced workers, in the case of the DiscoGEM corpus \citep{scholman-etal-2022-discogem}. GUM was annotated using RST as part of a Corpus Linguistics class, while DiscoGEM was annotated following the Penn Discourse Treebank \citep[PDTB,][]{prasad-etal-2008-penn,prasad-etal-2018-discourse} framework, using a method for crowdsourcing annotations introduced in \citet{yung-etal-2019-crowdsourcing}, and using a multi-label approach. 
The present work deviates from prior work in its focus on conversational data and the use of Segmented Discourse Representation Theory \citep[SDRT,][]{asher2003logics} alongside the STAC corpus \citep{asher-etal-2016-discourse} guidelines.

\section{Discourse relation annotation}

In this work, we focus on a subset of 19 dialogues from the Switchboard Corpus \cite{10.5555/1895550.1895693}.
This corpus contains informal language and has been the subject of study of numerous analyses of dialogue within linguistics \cite{jaeger2013alignment,Reitter2014AlignmentAT}.
In it, two strangers are presented with a topic (e.g., childcare) that they must discuss with each other, but the dialogues are otherwise not tightly constrained.
Annotating Switchboard will provide us with a more complete understanding of the use and generality of discourse relations across linguistic contexts and genres.

Following the annotation procedure in the STAC corpus \cite{asher-etal-2016-discourse}, we identified a subset of suitable elementary discourse units (EDUs) for annotation by parsing each turn into a dependency structure and included only those turns with at least two roots or verbs. 
Then, we segmented each of these turns into their respective EDUs.

Using these segmentations, we identified EDU candidates for discourse relations that were either within-turn (same speaker) or across two turns (different speakers, or the same speaker), where the two turns were adjacent in the case of different speakers, or only interrupted by one turn, in the case of same speaker. 
We provide a representative set of these pair types in Table \ref{table:relations} under the Explanation, Comment, and Result examples, respectively.

\subsection{Elementary Discourse Units}

Elementary discourse units (EDUs) are typically defined as non-overlapping text spans \citep{mann1987rhetorical}, which perform some basic discourse function \citep{asher2003logics}, typically at the level of clauses.
However, conversational EDUs may not necessarily contain a main verb (e.g., clarification questions: \say{Saginaw?}) or may be incomplete or interrupted (e.g., \say{and so--}). 
So, we define EDUs in Switchboard similarly to written text, with some modifications to account for variability due to spoken language. 
In particular, our modifications account for noise; non-linguistic communication (e.g., laughter); restarts; and disfluencies (e.g., \say{uh} or \say{um}).
Additionally, we use complex discourse units (CDUs), which are combinations of EDUs which function together as an argument to a DR \citep{asher2003logics}.

\begin{table*}[t]
\small
\begin{tabularx}{\textwidth}{r  X}
 \textbf{Relation} & \textbf{Discourse Units} \\ 
\hline
 Acknowledgement & A:  || \textit{it starts recording now.} || 
 
 B: || \textbf{Okay.} || \\ 
 Background & A: || \textit{I'm, we're originally from another state} || \textbf{and I know || in the state we were from that they did that t-, similar type thing.} || \\ 
 Clarification Question & A: || \textit{We live in the Saginaw area.} || 
 
 B: || \textbf{Saginaw?}|| \\ 
 Comment & B: || \textit{They seem to be having a real good response.} ||

A: || \textbf{That's pretty good.} || \\ 
 Continuation & A:  || \textit{I work off and on just temporarily} || \textbf{and usually find friends to babysit,} || \\ 
 Contrast & A: || \textit{I don't work, though,} || \textbf{but I used to work and,} || \\ 
Elaboration & A: || \textit{in the state we were from that they did that t-, similar type thing.} ||  \textbf{The city brought ought, || you know, || set tr-, separate trash cans || and you separated your stuff} || \\
Explanation & A: || \textit{and they discontinued them} || \textbf{because people were coming and dumping their trash in them.} || \\
Narration & A: || \textit{and you put it in there} || \textbf{and they took it,} || \\
Question-Answer Pair &  B: || \textit{Saginaw?} || 

A: || \textbf{Uh-huh.}|| \\
Result & B: || \textit{No, || I just, I noticed || it Iowa and other cities like that, it's a nickel per aluminum can.} ||

A: || Oh. ||

B: || \textbf{So you don't see too many thrown out around the || [laughter] || streets.} || \\
Other & None of the labels applies \\
\hline
\end{tabularx}
\caption{Representative discourse unit pairs for annotated discourse relations. The first argument to the discourse relation is shown in \textit{italics} and the second one in \textbf{bold}. $A$ and $B$ correspond to speakers, and double pipes (||) represent boundaries between elementary discourse units.}
\label{table:relations}
\end{table*}
\subsection{Relation categories}

Discourse relations (DRs) were selected from Segmented Discourse Representation theory (SDRT, \citealp{asher2003logics}), following the annotation manual for the STAC corpus \citep{asher2012manual}. 
11 out of 16 relation labels used in \citet{asher2012manual} were selected, based on a pilot annotation. 
We selected the most common relations in an attempt to minimize the number of choices presented to annotators, but the set is non-exhaustive. 
An "Other" category was added for cases in which none of the selected labels applied.
Table \ref{table:relations} shows the list of DRs together with representative examples.

\subsection{Annotators}
\label{Novices}
The present study recruited 114 students enrolled in a computational linguistics course grouped into 19 teams consisting of approximately 5 members who annotated the dyads.
Each team received a conversation for annotation. 
Annotations were performed individually, but groups then discussed their work and submitted a report as a team.
One team was excluded because they completed their annotations together and submitted a single set of labels.
Students were trained to identify discourse relations using a short quiz and live training with the instructor of the course.
Annotators were provided with guidelines to which they could refer back, 
and they had read and annotated the conversation in three previous tasks before annotating discourse relations, to ensure that they were familiar with the topics and speakers in each dyad. 

\subsection{Annotation procedure}
Annotators were presented with pairs representing either an EDU or CDU ($\pi_1$) and another EDU or CDU ($\pi_2$). 
Annotators were shown two spans of text $\pi_1$ and $\pi_2$ with $\pi_1$ presented in italics and $\pi_2$ presented in bold face font in the annotation software Prodigy \citep{montani2018prodigy}, with two preceding and two subsequent turns for context. 
Annotators were asked to determine the relation between $\pi_1$ and $\pi_2$ from a list of the DR categories in Table \ref{table:relations}. If annotators thought that no relation was present, they were told to reject the item and move on to the next pair.
Critically for our research question, annotators could mark several relations for a pair of EDUs simultaneously.
In addition to labeling discourse relations, annotators were also asked to provide a confidence rating on a scale from 1-5, but we leave these analyses for future work.
In total, each annotator provided judgments for an average of 25 EDU pairs across 464 total pairs. 

In the next section, we test whether annotators show greater uncertainty about discourse relations in different discourse contexts. We analyze the distribution of their labels to assess whether discourse relations in conversation vary in their contexts of use.

\section{Uncertainty in the annotation of discourse relations}

\begin{figure*}[h]
    \centering
    \includegraphics[width=.825\textwidth]{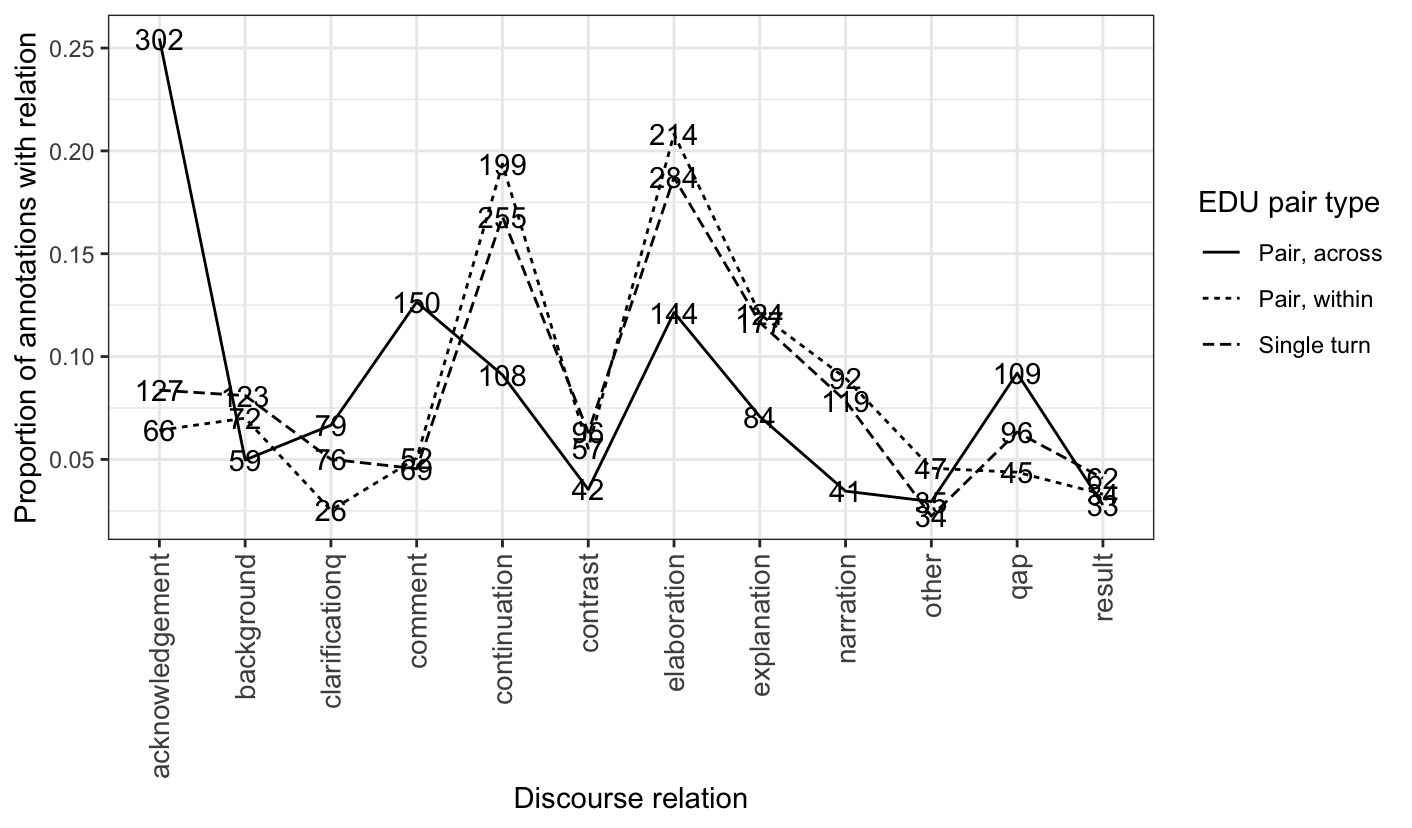}
    \caption{Distribution of discourse relations across three EDU pair types. y axis represents proportions of EDU pairs with a given label; numbers represent the count of a label within a discourse context category.}
    \label{fig:frequencies}
\end{figure*}

Different EDU pairs in the present annotation task were drawn either from the same turn, or across turns but within or across speakers. 
Thus, we can assess how much discourse relations vary by the placement of an utterance in a dialogue.
Given the complex dynamics in dialogue, we expect to find significant differences in discourse relation use across different discourse contexts.
We visualize the distribution of the relations in Figure \ref{fig:frequencies}.

Annotators generally selected more discourse relations per EDU pair in the single-turn case, with an average of 8.16 relations per team or 1.60 per annotator.
When EDUs spanned turns within a single speaker, groups selected significantly fewer relations (average = 7.29, $t(302) = -2.16$, $p < .05$).
Groups likewise selected even fewer relations for EDUs between two speakers (average = 6.51, $t(314) = -2.54$, $p < .05$).
On its face, this pattern appears surprising, because it suggests that annotators find more relations appropriate for single-speaker productions.
However, an alternative interpretation of these results is that annotators may instead have been uncertain about the distinctions between the different discourse relations.
This second interpretation is corroborated by post-hoc poll data from 35 annotators, of whom 32 (91.4\%) stated that the selection of discourse relations was best suited to annotating cross-speaker EDU pairs.
Future work will require recruiting greater numbers of annotators to be able to distinguish between these two hypotheses.

\subsection{Inter-annotator agreement}

We computed measures of inter-annotator agreement for multilabel tasks using \citet{marchal-etal-2022-establishing}.
This approach uses bootstrap sampling to estimate the chance frequencies of DRs in a multi-label dataset to provide a baseline for agreement between annotators.

We summarize the results of this analysis in Table \ref{table:agreement}. Following \citet{marchal-etal-2022-establishing}, we computed observed, expected and adjusted agreement for six measures. Soft-match agreement uses the intersection of labels selected by two annotators; boot-match corrects the expected agreement by using the bootstrapping method (as opposed to ignoring non-intersecting labels); augmented kappa uses DR labels weighted according to the number of labels annotated for each item; precision and recall are calculated as the proportion of intersecting DR labels over the set of labels selected by the first and second annotator, respectively; F1 is the usual harmonic mean between precision and recall. 

Both observed and adjusted agreement metrics were well above chance using the bootstrapping method proposed by \citet{marchal-etal-2022-establishing}. Agreement is in general modest \cite{Landis1977TheMO}, which may be partly due to the challenging nature of the DRs annotation task \cite{spooren2010coding}, and partly due to annotators' uncertainty on DR labels across different context types.

\begin{table}[t]
\small
\begin{center}
    \begin{tabular}{lrrr}
    \hline
        &       observed  & expected     & adjusted kappa \\
        \hline
    \texttt{soft-match}  &   0.43 & 0.11 & 0.36 \\ 
    \texttt{augmented}       & 0.27 & 0.11 & 0.18 \\
        \texttt{boot-match}     & 0.43 & 0.21 & 0.27\\
    \texttt{boot-rec.}    & 0.33 & 0.14 & 0.22\\
        \texttt{boot-prec.} & 0.36 & 0.17 & 0.23\\
            \texttt{boot-F1}      &  0.32 & 0.13 & 0.21\\
    \hline
    \end{tabular}
\end{center}
\caption{Outputs of \citet{marchal-etal-2022-establishing} inter-annotator agreement analysis.}
\label{table:agreement}
\end{table}

\subsection{Predicting relation selection}

\begin{table}[]
\begin{center}
\small
\begin{tabular}{rrrr}
 \textbf{Relation}              & \textbf{Intercept} & \textbf{\shortstack{Different\\ speaker}} & \textbf{\shortstack{Within\\ turn}} \\
 \hline
Background     & -2.73  & 1.96                & 0.24            \\
Clarification Q. & 0.02  & 0.41               & -0.43            \\
Comment        & -1.69  & 0.84                & 0.46            \\
Continuation   & -1.60  & 2.35               & 0.31            \\
Contrast       & -2.23   & 2.07               & 0.17            \\
Elaboration    & -0.78  & 1.95               & 0.13            \\
Explanation    & -1.40  & 2.00                & 0.09            \\
Narration      & -3.29  & 2.68               & 0.53            \\
Other          & -3.54  & 2.20               & 1.12            \\
Q-A Pair            & -0.81  & 0.64               & -0.01            \\
Result         & -2.35  & 1.63                & -0.00   \\        
\hline
\end{tabular}
\end{center}
\caption{Coefficient estimates from a multiclass logistic regression predicting each annotation label.}
\label{table:multinom_analyses}
\end{table}

We use a model comparison approach to understand the contributions of discourse context (within/across speakers and within/across turns) to relation annotation by first constructing a null model that estimates the base rates of each discourse relation.
Then, we constructed a multiclass logistic regression model containing the discourse context variables of interest, which significantly improved fit to the annotation data ($X^{2}(22) = 447.98$, $p < .001$).
This improvement in fit suggests that the distribution of discourse relations that are identified by annotators is distinct across contexts.
Adding the annotator group/topic also significantly improved fit beyond the model containing the contextual variables alone  ($X^{2}(198) = 900.06$ $p < .001$).
We summarize the results of this final model in Table \ref{table:multinom_analyses}.\footnote{Due to the multilabel nature of the annotation task and the one-versus-rest training for the multiclass model, coefficients for each DR are not independent, were not estimated jointly, and should be interpreted broadly as representing separate logistic regressions.}

An informal evaluation of the coefficients suggests that discourse relations are not uniformly distributed across contexts.
Intuitively, Acknowledgements, Clarification Questions, Comments, and Question-Answer Pairs are more likely across speakers than within.
Additionally, Continuations, Elaborations, Explanations, and Narrations are more likely to occur within a single speaker.
The pattern of results is more unclear when comparing EDUs that are produced by a single speaker but which occur either within or across turns.
For example, relations such as Clarification Questions are less likely to occur within a turn than across turns.

\subsection{Classifier for relations}

To validate the quality of the annotations, we built a model to classify EDU pairs into discourse relations. 
We reasoned that if annotators are following the guidelines and use information about the EDU pairs, then a classifier should be able to predict DR labels.
We encoded the first EDU or CDU ($\pi_1$) and the second ($\pi_2$) as the two \say{sentences} in the next sentence prediction architecture of BERT \cite{devlin-etal-2019-bert}.
This enables the classifier to represent the $\pi_1$ and $\pi_2$ components somewhat separately.

We built a classifier head trained on the resulting embeddings without fine-tuning to predict each individual annotator label.
We chose to model each annotator label individually to learn agreement/majority class implicitly because prior studies have shown that this improves generalization \cite{yung-etal-2022-label}.
We use a leave-one-conversation-out training procedure, in which we test a ridge regression classifier on all of the annotations from a single conversation while we train it on all other annotations across the other conversations.
This ensures minimal memorization of specific turns within a conversation, which is critical given our multilabel annotation approach.

Strict annotation-level accuracy to predict each selected label from all annotators was quite poor, with macro average precision at $.21$, recall at $.19$, and F1 at $.19$.
However, recall was substantially higher when considering whether the top guess belonged to the set of all labels provided by annotators, at $.76$ overall and $.71$ averaged by group.

To quantify the uncertainty of the annotators across different contexts, we leverage the classifier to produce a label distribution for a given $(\pi_1, \pi_2)$ pair.
We then compute the cross-entropy between the model's predictions and annotators' gold label distributions, collapsing across all annotations for an EDU pair.
Overall, cross entropy between model predictions and annotator labels was highest for the single-turn case, with (mean $ = 0.43$), but lowest for EDUs between two speakers (mean $ = 0.38$), suggesting greater uncertainty in label assignment.

\section{Discussion}

In two experiments, we demonstrated that novice DR label annotations in a single turn are more difficult than across turns.
We showed that including discourse context (within/across speaker and within/across turn) to a logistic regression model significantly improves fit to our annotation data. 
A classifier trained to predict DR labels from embeddings of $(\pi_1, \pi_2)$ pairs showed modest success for recall of any of the annotations, but poor precision and recall overall. 
A comparison of this classifier's predictions and annotators' gold label distributions revealed greater uncertainty for the annotation of discourse relations within a single turn.

These results demonstrate that different conversational contexts are associated with different distributions of discourse relations. The uncertainty of choice of discourse relations within a turn may be due to several factors. DRs that typically occur across adjacent turns and across speakers (e.g., Acknowledgements) might have clearer signals. At the same time, DRs that occur more frequently within speakers, and, in particular, within a turn, might be more ambiguous, or might co-occur with other relations. More work is necessary to disentangle uncertainty about the identity of the best fit relation from whether multiple relations are appropriate.

\section*{Limitations}
The current work is limited by the size of the dataset and the nature of spontaneous conversation. While the discourse relations proposed as part of this work were selected to be general and build on categories from the literature, the list is not exhaustive and it is likely that these relations may be culturally, linguistically, and situationally specific.
Future work in this area should validate the generality of the discourse relation system used in this work.

The selection of EDUs and CDUs for annotation is also non-exhaustive; additional segments could be included in future work. 

Annotation quality is also a practical limitation. 
Annotation for discourse relations typically results in low-agreement data, even among expert annotators (e.g., DiscoGEM; \citealp{scholman-etal-2022-discogem}).
Even though our research questions focus on this disagreement as a positive, other researchers may require greater numbers of annotations in order to obtain a gold label.

\section*{Ethics Statement}
We are not aware of ethical issues associated with the texts used in this work. Students participated in the annotation task as part of course credit but annotation decisions were not associated with their performance in the course. 

\section*{Acknowledgements}
We would like to thank Jürgen Bohnemeyer and three anonymous reviewers for feedback on a previous version of this paper.
\bibliographystyle{acl_natbib}

\bibliography{anthology,custom}

\begin{thebibliography}{28}
\expandafter\ifx\csname natexlab\endcsname\relax\def\natexlab#1{#1}\fi

\bibitem[{Afantenos et~al.(2015)Afantenos, Kow, Asher, and
  Perret}]{afantenos-etal-2015-discourse}
Stergos Afantenos, Eric Kow, Nicholas Asher, and J{\'e}r{\'e}my Perret. 2015.
\newblock \href {https://doi.org/10.18653/v1/D15-1109} {Discourse parsing for
  multi-party chat dialogues}.
\newblock In \emph{Proceedings of the 2015 Conference on Empirical Methods in
  Natural Language Processing}, pages 928--937, Lisbon, Portugal. Association
  for Computational Linguistics.

\bibitem[{Asher et~al.(2016)Asher, Hunter, Morey, Farah, and
  Afantenos}]{asher-etal-2016-discourse}
Nicholas Asher, Julie Hunter, Mathieu Morey, Benamara Farah, and Stergos
  Afantenos. 2016.
\newblock \href {https://aclanthology.org/L16-1432} {Discourse structure and
  dialogue acts in multiparty dialogue: the {STAC} corpus}.
\newblock In \emph{Proceedings of the Tenth International Conference on
  Language Resources and Evaluation ({LREC}'16)}, pages 2721--2727,
  Portoro{\v{z}}, Slovenia. European Language Resources Association (ELRA).

\bibitem[{Asher and Lascarides(2003)}]{asher2003logics}
Nicholas Asher and Alex Lascarides. 2003.
\newblock \emph{Logics of conversation}.
\newblock Cambridge University Press.

\bibitem[{Asher et~al.(2012)Asher, Popescu, Muller, Afantenos, Cadilhac,
  Benamara, Vieu, and Denis}]{asher2012manual}
Nicholas Asher, Vladimir Popescu, Philippe Muller, Stergos Afantenos, Anais
  Cadilhac, Farah Benamara, Laure Vieu, and Pascal Denis. 2012.
\newblock Manual for the analysis of settlers data.
\newblock \emph{Strategic Conversation (STAC). Universit{\'e} Paul Sabatier}.

\bibitem[{Carlson et~al.(2003)Carlson, Marcu, and
  Okurowski}]{carlson2003building}
Lynn Carlson, Daniel Marcu, and Mary~Ellen Okurowski. 2003.
\newblock Building a discourse-tagged corpus in the framework of rhetorical
  structure theory.
\newblock In \emph{Current and new directions in discourse and dialogue}, pages
  85--112. Springer.

\bibitem[{Crible and Cuenca(2017)}]{crible2017discourse}
Ludivine Crible and Maria-Josep Cuenca. 2017.
\newblock Discourse markers in speech: distinctive features and corpus
  annotation.
\newblock \emph{Dialogue and Discourse}, 8(2):149--166.

\bibitem[{Devlin et~al.(2019)Devlin, Chang, Lee, and
  Toutanova}]{devlin-etal-2019-bert}
Jacob Devlin, Ming-Wei Chang, Kenton Lee, and Kristina Toutanova. 2019.
\newblock \href {https://doi.org/10.18653/v1/N19-1423} {{BERT}: Pre-training of
  deep bidirectional transformers for language understanding}.
\newblock In \emph{Proceedings of the 2019 Conference of the North {A}merican
  Chapter of the Association for Computational Linguistics: Human Language
  Technologies, Volume 1 (Long and Short Papers)}, pages 4171--4186,
  Minneapolis, Minnesota. Association for Computational Linguistics.

\bibitem[{Godfrey et~al.(1992)Godfrey, Holliman, and
  McDaniel}]{10.5555/1895550.1895693}
John~J. Godfrey, Edward~C. Holliman, and Jane McDaniel. 1992.
\newblock Switchboard: Telephone speech corpus for research and development.
\newblock In \emph{Proceedings of the 1992 IEEE International Conference on
  Acoustics, Speech and Signal Processing - Volume 1}, ICASSP'92, page
  517–520, USA. IEEE Computer Society.

\bibitem[{Jaeger and Snider(2013)}]{jaeger2013alignment}
T~Florian Jaeger and Neal~E Snider. 2013.
\newblock Alignment as a consequence of expectation adaptation: Syntactic
  priming is affected by the prime’s prediction error given both prior and
  recent experience.
\newblock \emph{Cognition}, 127(1):57--83.

\bibitem[{Jansen et~al.(2014)Jansen, Surdeanu, and Clark}]{jansen2014discourse}
Peter Jansen, Mihai Surdeanu, and Peter Clark. 2014.
\newblock Discourse complements lexical semantics for non-factoid answer
  reranking.
\newblock In \emph{Proceedings of the 52nd Annual Meeting of the Association
  for Computational Linguistics (Volume 1: Long Papers)}, pages 977--986.

\bibitem[{Kraus and Feuerriegel(2019)}]{kraus2019sentiment}
Mathias Kraus and Stefan Feuerriegel. 2019.
\newblock Sentiment analysis based on rhetorical structure theory: Learning
  deep neural networks from discourse trees.
\newblock \emph{Expert Systems with Applications}, 118:65--79.

\bibitem[{Landis and Koch(1977)}]{Landis1977TheMO}
J~Richard Landis and Gary~G. Koch. 1977.
\newblock The measurement of observer agreement for categorical data.
\newblock \emph{Biometrics}, 33 1:159--74.

\bibitem[{Levinson and Torreira(2015)}]{Levinson2015TimingIT}
Stephen~C. Levinson and Francisco Torreira. 2015.
\newblock Timing in turn-taking and its implications for processing models of
  language.
\newblock \emph{Frontiers in Psychology}, 6.

\bibitem[{Liu and Chen(2019)}]{liu-chen-2019-exploiting}
Zhengyuan Liu and Nancy Chen. 2019.
\newblock \href {https://doi.org/10.18653/v1/D19-5415} {Exploiting
  discourse-level segmentation for extractive summarization}.
\newblock In \emph{Proceedings of the 2nd Workshop on New Frontiers in
  Summarization}, pages 116--121, Hong Kong, China. Association for
  Computational Linguistics.

\bibitem[{Mann and Thompson(1987)}]{mann1987rhetorical}
William~C Mann and Sandra~A Thompson. 1987.
\newblock \emph{Rhetorical {S}tructure {T}heory: A theory of text
  organization}.
\newblock University of Southern California, Information Sciences Institute Los
  Angeles.

\bibitem[{Marchal et~al.(2022)Marchal, Scholman, Yung, and
  Demberg}]{marchal-etal-2022-establishing}
Marian Marchal, Merel Scholman, Frances Yung, and Vera Demberg. 2022.
\newblock \href {https://aclanthology.org/2022.coling-1.322} {Establishing
  annotation quality in multi-label annotations}.
\newblock In \emph{Proceedings of the 29th International Conference on
  Computational Linguistics}, pages 3659--3668, Gyeongju, Republic of Korea.
  International Committee on Computational Linguistics.

\bibitem[{Meyer and Popescu-Belis(2012)}]{meyer2012using}
Thomas Meyer and Andrei Popescu-Belis. 2012.
\newblock Using sense-labeled discourse connectives for statistical machine
  translation.
\newblock In \emph{EACL 2012: Proceedings of the Joint Workshop on Exploiting
  Synergies between Information Retrieval and Machine Translation (ESIRMT) and
  Hybrid Approaches to Machine Translation (HyTra)}, CONF, pages 129--138.

\bibitem[{Montani and Honnibal(2018)}]{montani2018prodigy}
Ines Montani and Matthew Honnibal. 2018.
\newblock Prodigy: A new annotation tool for radically efficient machine
  teaching.
\newblock \emph{Artificial Intelligence}.

\bibitem[{Prasad et~al.(2008)Prasad, Dinesh, Lee, Miltsakaki, Robaldo, Joshi,
  and Webber}]{prasad-etal-2008-penn}
Rashmi Prasad, Nikhil Dinesh, Alan Lee, Eleni Miltsakaki, Livio Robaldo,
  Aravind Joshi, and Bonnie Webber. 2008.
\newblock \href
  {http://www.lrec-conf.org/proceedings/lrec2008/pdf/754_paper.pdf} {The {P}enn
  {D}iscourse {T}ree{B}ank 2.0.}
\newblock In \emph{Proceedings of the Sixth International Conference on
  Language Resources and Evaluation ({LREC}'08)}, Marrakech, Morocco. European
  Language Resources Association (ELRA).

\bibitem[{Prasad et~al.(2018)Prasad, Webber, and
  Lee}]{prasad-etal-2018-discourse}
Rashmi Prasad, Bonnie Webber, and Alan Lee. 2018.
\newblock \href {https://aclanthology.org/W18-4710} {Discourse annotation in
  the {PDTB}: The next generation}.
\newblock In \emph{Proceedings 14th Joint {ACL} - {ISO} Workshop on
  Interoperable Semantic Annotation}, pages 87--97, Santa Fe, New Mexico, USA.
  Association for Computational Linguistics.

\bibitem[{Reitter and Moore(2014)}]{Reitter2014AlignmentAT}
D.~Reitter and Johanna~D. Moore. 2014.
\newblock Alignment and task success in spoken dialogue.
\newblock \emph{Journal of Memory and Language}, 76:29--46.

\bibitem[{Roberts(2012)}]{roberts2012information}
Craige Roberts. 2012.
\newblock Information structure: Towards an integrated formal theory of
  pragmatics.
\newblock \emph{Semantics and pragmatics}, 5:6--1.

\bibitem[{Scholman et~al.(2022)Scholman, Dong, Yung, and
  Demberg}]{scholman-etal-2022-discogem}
Merel Scholman, Tianai Dong, Frances Yung, and Vera Demberg. 2022.
\newblock \href {https://aclanthology.org/2022.lrec-1.351} {{D}isco{G}e{M}: A
  crowdsourced corpus of genre-mixed implicit discourse relations}.
\newblock In \emph{Proceedings of the Thirteenth Language Resources and
  Evaluation Conference}, pages 3281--3290, Marseille, France. European
  Language Resources Association.

\bibitem[{Spooren and Degand(2010)}]{spooren2010coding}
Wilbert Spooren and Liesbeth Degand. 2010.
\newblock Coding coherence relations: Reliability and validity.
\newblock \emph{Corpus Linguistics and Linguistic Theory}, 6(2):241--266.

\bibitem[{Tonelli et~al.(2010)Tonelli, Riccardi, Prasad, and
  Joshi}]{tonelli-etal-2010-annotation}
Sara Tonelli, Giuseppe Riccardi, Rashmi Prasad, and Aravind Joshi. 2010.
\newblock \href
  {http://www.lrec-conf.org/proceedings/lrec2010/pdf/184_Paper.pdf} {Annotation
  of discourse relations for conversational spoken dialogs}.
\newblock In \emph{Proceedings of the Seventh International Conference on
  Language Resources and Evaluation ({LREC}'10)}, Valletta, Malta. European
  Language Resources Association (ELRA).

\bibitem[{Yung et~al.(2022)Yung, Anuranjana, Scholman, and
  Demberg}]{yung-etal-2022-label}
Frances Yung, Kaveri Anuranjana, Merel Scholman, and Vera Demberg. 2022.
\newblock \href {https://aclanthology.org/2022.codi-1.7} {Label distributions
  help implicit discourse relation classification}.
\newblock In \emph{Proceedings of the 3rd Workshop on Computational Approaches
  to Discourse}, pages 48--53, Gyeongju, Republic of Korea and Online.
  International Conference on Computational Linguistics.

\bibitem[{Yung et~al.(2019)Yung, Demberg, and
  Scholman}]{yung-etal-2019-crowdsourcing}
Frances Yung, Vera Demberg, and Merel Scholman. 2019.
\newblock \href {https://doi.org/10.18653/v1/W19-4003} {Crowdsourcing discourse
  relation annotations by a two-step connective insertion task}.
\newblock In \emph{Proceedings of the 13th Linguistic Annotation Workshop},
  pages 16--25, Florence, Italy. Association for Computational Linguistics.

\bibitem[{Zeldes(2017)}]{zeldes2017gum}
Amir Zeldes. 2017.
\newblock The {GUM} corpus: Creating multilayer resources in the classroom.
\newblock \emph{Language Resources and Evaluation}, 51(3):581--612.

\end{thebibliography}

\end{document}